\documentclass[10pt, a4paper]{article}

\usepackage[final]{lrec2026} 

\usepackage{float}
\usepackage{amsmath}
\usepackage{multirow}

\usepackage{array}

\newcommand{\mainstream}[0]{\texttt{MAINSTREAM}}

\newcommand{\ppn}[0]{\texttt{PPN}}

\usepackage{subfigure}
\usepackage{xurl}      
\usepackage[table]{xcolor}
\usepackage{hyperref}

\title{Reliable News or Propagandist News?
A Neurosymbolic Model Using Genre, Topic, and Persuasion Techniques to Improve Robustness in Classification}

\name{Géraud Faye$^{1,2}$, {\bf \large Benjamin Icard$^{3}$}, {\bf \large Morgane Casanova$^{4}$},\\  {\bf \large Guillaume Gadek$^{1}$}, {\bf \large Guillaume Gravier$^{4}$}, {\bf \large Wassila Ouerdane$^{2}$},\\  {\bf \large Céline Hudelot$^{2}$},  {\bf \large Sylvain Gatepaille$^{1}$}, {\bf \large Paul \'Egr\'e$^{5}$}}

\address{\vspace{0.1in} \\    
          $^{1}$Airbus Defence and Space, France\\
          $^{2}$Université Paris-Saclay, CentraleSupélec, MICS, France\\
          $^{3}$LIP6, Sorbonne Université, CNRS, France \\
          $^{4}$Université de Rennes, CNRS, Inria, IRISA, France\\
          $^{5}$IRL Crossing, CNRS, Australia
  \vspace{0.2in}        
}

\abstract{Among news disorders, propagandist news are particularly insidious, because they tend to mix oriented messages with factual reports intended to look like reliable news. To detect propaganda, extant approaches based on Language Models such as BERT are promising but often overfit their training datasets, due to biases in data collection. To enhance classification robustness and improve generalization to new sources, we propose a neurosymbolic approach combining non-contextual text embeddings (fastText) with symbolic conceptual features such as genre, topic, and persuasion techniques. Results show improvements over equivalent text-only methods, and ablation studies as well as explainability analyses confirm the benefits of the added features.
 \\ \newline }

\begin{document}

\maketitleabstract

\section{Introduction}
\label{sec:intro}

Recent years have seen a sharp increase in online news manipulation, driven by renewed international tensions, as documented in Europe by various intelligence offices (VIGINUM\footnote{\url{https://www.sgdsn.gouv.fr/notre-organisation/composantes/service-de-vigilance-et-protection-contre-les-ingerences-numeriques?}} in France, ZEAM\footnote{\url{https://www.bmi.bund.de/SharedDocs/schwerpunkte/EN/disinformation-election/zeam-artikel-en.html?}} in Germany) and monitoring organizations (viz. EU DisinfoLab\footnote{\url{https://www.disinfo.eu}}). Such manipulation of information, which we may refer to as ``news disorder'' (adapting the terminology of \citealt{wardlederakhshan2017informationdisorder}), is often orchestrated through press-like websites that mimic journalistic conventions and disseminate targeted narratives to shape opinion (aka. ``pseudo-news'', see \citealt{faye-etal-2024-exposing}). This is concerning since such content is widely shared on social media, quickly reaching large audiences.\footnote{In September 2025, Pew Research Center estimates that 53\% of U.S. adults used social media as a news source: \url{https://www.pewresearch.org/journalism/fact-sheet/social-media-and-news-fact-sheet/}.}

Automated detection of news disorders has advanced in various directions, including media bias \cite{hamborg2019mediabias}, misinformation via automated fact verification \cite{thorne-etal-2018-fever}, fake news \cite{10.1145/3395046,Hu2025OverviewFakeNewsDetection}, and rumors \cite{shu2020fakenewsnet}. However, these methods often transfer poorly across other types of news disorder. In particular, they tend to break down on influence operations and propaganda, which hinge on context-dependent framing of genres and topics, and on several specific persuasion techniques \cite{da-san-martino-etal-2019-fine,dasanmartino2020surveypropaganda}.


In this paper, we focus on the identification of propagandist news, which are particularly insidious because they tend to smuggle in oriented messages with factual reports intended to look like reliable news. For this detection task, LLMs show high performance on specific datasets, but suggestive of a potential overfit. To increase robustness in classification, we argue that hybrid approaches, which have been effective for other news disorders (e.g., \citealp{baly-etal-2018-predicting,thorne-etal-2018-fever,ruchansky2017csi,ma-etal-2017-detect}), can improve the detection of influence operations and propaganda in news. Drawing on existing corpora of transparent news versus propagandist news \citep{faye-etal-2024-exposing}, we present a neurosymbolic detection model that combines static vector embeddings (fastText) with additional features that include genre, topic, and persuasion techniques. We provide evidence, using biased splits of the training, validation, and test tests, that the incorporation of these features improves performance over text-only methods. 



Section~\ref{sec:sota} reviews related work on fake news detection and on the challenges posed by propaganda. Section~\ref{sec:data} introduces two existing datasets that include reliable articles and then propaganda articles, to identify cross-corpus differences and features characteristic of propaganda.
Building on these observations, Section~\ref{sec:detection} introduces a neurosymbolic model that merges dense text embeddings with interpretable conceptual features automatically extracted from the texts. Section~\ref{sec:results} compares the performance of our hybrid method with that of a text-only benchmark; it evaluates its robustness relative to different partitions of the training/valid/test tests, and it uses ablation studies and explainability analyses to validate the method. Section~\ref{sec:conclusion} discusses the findings and outlines directions for future work.


\section{Related work}
\label{sec:sota}

The detection of information manipulation and news disorders is a broad domain that involves identifying different types of problematic content. The broadest category for misinformation is that of \textit{fake news}, often defined as false or biased information, produced with an agenda or by negligence \cite{baptista2022working}. Other varieties of news disorder include \textit{rumors}, typically news reporting information that is hardly verifiable when published, intending to capture public attention. A specific variety of fake news is \textit{propaganda}, namely partisan information seeking to set a narrative in order to debunk an enemy and glorify a state or organization.


{Fake news} detection~\cite{Hu_Wei_Zhao_Wu_2022,10.1145/3395046} is a popular area of research. It can be detected using transformer-based models~\cite{DBLP:journals/corr/abs-2104-06952}, or using linguistic features and web markup features~\cite{10.1145/3308560.3316739}, or combining linguistic and knowledge features~\cite{9791232,guelorget_combining}. 

Fake news can also be easily detected using social media propagation patterns~\cite{amila_propagation2vec,davoudi_dss}, reaching in some cases more than 99\% accuracy on four-day propagation data. However, in the current context, content-based fake news detection is more relevant, as a four-day delay is too long for effective detection. 

Other approaches rely on user reports~\cite{10.1145/3184558.3188722} and are designed to remain effective even when the majority of users engage in malicious reporting behavior. Various datasets exist for this subtask, often annotated by journalists, such as PolitiFact~\cite{shu2020fakenewsnet} and \citet{Horne_Adali_2017} on US politics, CoAID~\cite{coaid} on Covid-19, as well as LIAR~\cite{wang-2017-liar}, MultiFC~\cite{augenstein-etal-2019-multifc} or MuMIN~\cite{nielsen_2022_mumin} on diverse topics. In the case of rumors, related datasets for this task are Fakenewsnet~\cite{shu2020fakenewsnet}, relying on labels produced by PolitiFact and GossipCop, and PHEME~\cite{kochkina_2018_pheme}.

A method put forward to analyze the news and to track whether they count as reliable or fake is \textit{stance classification}~\cite{Riedel2017ASB}. In one version of the task, the goal is to assign a claim-evidence pair to one of three categories: the evidence either supports the claim, contradicts it, or fails to provide sufficient evidence. 
The primary dataset for this task is FEVER~\cite{thorne-etal-2018-fever}, containing more than 300,000 facts. This task could help detect misinformation based on the content to be checked and a small collection of related evidence, making this task also close to fact-checking. The method has also been used to detect propaganda \cite{hanley2025tracking}.




Propaganda differs from other forms of news disorder by explicitly mimicking news articles and relying on frames and persuasion techniques \cite{dasanmartino2020surveypropaganda}. \citealp{barroncedeno2019proppy} computes a propaganda score, defined as the estimated likelihood that an article contains propagandistic mechanisms, using engineered stylistic and lexical features such as readability, lexical richness, and TF-IDF n-grams, whereas \citealp{da-san-martino-etal-2019-fine} introduces a fragment-level annotation scheme in which expert annotators mark the exact spans in news articles that realize propaganda and label each marked span with one of 18 propaganda persuasion techniques. This formulation, combining span identification with technique classification, is benchmarked in NLP4IF-2019 \cite{dasanmartino2019nlp4if} and SemEval-2020 Task~11 \cite{dasanmartino2020semeval11}. 
 
\citet{faye-etal-2024-exposing} compare human annotations with model predictions for multi-label propaganda analysis of press articles and systematically evaluates which stylistic cues explain performance. In particular, they show that feature sets targeting vagueness and subjectivity, together with syntactic and lexical cues, can achieve performance comparable to RoBERTa, while making the textual correlates of predictions more explicit.


Classical content-only approaches for news disorder detection often overfit to dataset or domain artifacts and therefore generalize poorly across outlets, genres, topics, and types of news disorder \cite{suprem2022generalizability,silva2021crossdomain,pan2023zeroshot,krieger2022domain}. To improve robustness, neurosymbolic hybrid approaches pair neural representations of news articles with structured, complementary signals such as external evidence, source metadata, lexicon and stylistic indicators, or multimodal cues. This strategy has been effective for misinformation through evidence-based verification and debunking \cite{thorne-etal-2018-fever,popat-etal-2018-declare}, for fake news by incorporating social-context and multimodal signals \cite{ruchansky2017csi,wang2018eann}, and for media bias by combining article text with source-level features and distant supervision \cite{baly-etal-2018-predicting,spinde-etal-2021-neural-media}.

In the wake of these approaches, here we present a neurosymbolic model for propaganda detection grounded in stylistic features and observable symbolic cues, motivated by differences reported across previously published news corpora. We first conduct a comparative analysis of propaganda and mainstream news, then leverage the observed contrasts to design and evaluate our feature-based model using different organizations of our dataset to aim for more robustness.

\section{Datasets used}
\label{sec:data}

\subsection{Two corpora}


In the rest of this paper, we exploit two datasets recently presented in \citet{faye-etal-2024-exposing}, which respectively consist of a corpus of propagandist pseudo-news articles (\ppn) and a corpus of reliable articles from the mainstream press (\mainstream).

\begin{itemize}
    \item \texttt{PPN}\footnote{\url{https://github.com/hybrinfox/ppn}}~\cite{faye-etal-2024-exposing}, for Propagandist Pseudo-News, is a collection of 12,427 articles from sources identified as propaganda outlets by the expert organizations NewsGuard and VIGINUM.\footnote{\url{https://www.sgdsn.gouv.fr/publications/maj-19062023-rrn-une-campagne-numerique-de-manipulation-de-linformation-complexe-et}} The five sources were created after the Russian invasion of Ukraine in February 2022 and contain propagandist news in 9 different languages (Arabic, Chinese, English, French, German, Italian, Russian, Spanish and Ukrainian). 
    
    \item \mainstream\ is a corpus of French and English articles, this time of regular news coming from established newspapers, and used as a control for the analysis of the \ppn\ corpus. The \mainstream\ articles were selected based on publication dates and on keywords related to the Ukraine conflict. \mainstream\ consists of
    1,004 English articles and 1,367 French articles from 11 and 5 sources, respectively.
\end{itemize}

While both datasets were introduced in \citet{faye-etal-2024-exposing}, only a small portion was analyzed in the context of an annotation experiment (100 in French split across both sources). Here, we analyze the whole corpus in full and using different methods.\footnote{Because of copyright issues, the text is not redistributed, but direct links to web pages are provided with the corresponding annotations.}


\subsection{Review of published observations}

\begin{figure}
    \centering
    \includegraphics[width=\linewidth]{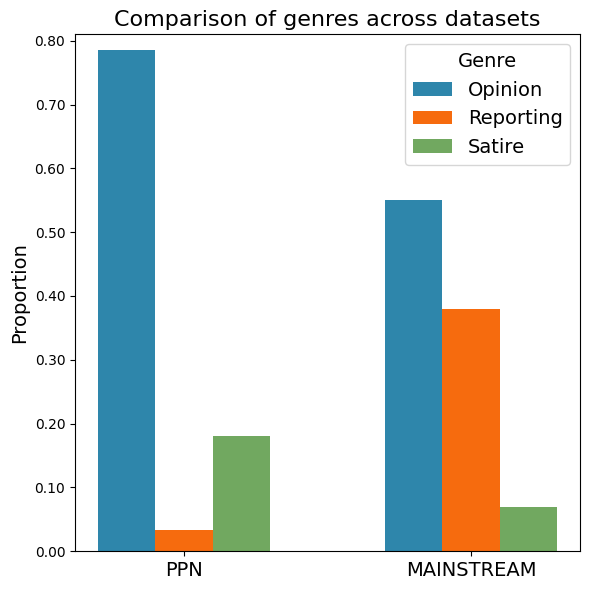}
    \caption{News genre distribution on the two corpora.}
    \label{fig:genres}
\end{figure}

Before presenting new analyses, we briefly summarize the main results of previous analyzes.

Firstly, a French subset of \texttt{PPN} and \texttt{MAINSTREAM} was manually annotated using 11 labels, adapted from a previous annotation experiment.\footnote{\label{fn:obsinfox}See the OBSINFOX dataset, \url{https://github.com/obs-info/obsinfox} and its analysis in \citet{icard-etal-2024-multi-label}. The focus on French was to take advantage of the native competence of the annotators.} The labels included ``vague'', ``subjective'', ``exaggeration'', ``pejorative'', ``descriptive'', ``propaganda'', ``satirical'', ``dishonest title'', ``adequate sources'', ``false information'', and ``fake news''.\footnote{The difference between ``fake news'' and ``false information'' was that in order to be ascribed the former label, an article had to contain ``at least one false information''. The definition of ``fake news'' was deliberately left up to the annotators in the experiment, see \cite{faye-etal-2024-exposing} for details.} 

Mainstream articles were found to be generally descriptive (close to 90\%) and to adequately cite their sources (above 80\%). They received low scores for ``subjective'' (20\%) and ``vague'' (10\%), and hardly any of the other labels, including no ascription of ``false information'' (0\%). By contrast, propaganda articles were labeled as descriptive for only 60\%, and they were overwhelmingly labeled as manifesting subjectivity (70\%), with up to 30\% of ``false information'' and over 40\% of ``fake news''.

Secondly, Large Language Models fine-tuned for the task, such as RoBERTa-base~\cite{roberta}, were trained and tested on the whole English corpus. The LLMs were found to distinguish with very high performance between propaganda and regular articles (99.7\% of test accuracy). Other models, such as TF-IDF~\cite{tfidf}, gave a similar high performance (98.5\%), and also evidenced lexical differences between the two English corpora. 

In both cases, however, this very high performance raises suspicion, since it can be a sign that the methods won't transfer to new datasets. While we do not see direct evidence of overfitting in the comparison between our training, validation, and test tests, such high accuracy points to the risk of a lack of robustness regarding other datasets. 
Furthermore, it is also possible that articles from the \texttt{PPN} dataset are partially written by AI, adding further biases to the model, leading to more overfitting on machine-written misinformation~\cite{su2023fakenewsdetectorsbiased}.

\section{Genres, topics, and persuasion techniques}

To get a more general comparison of the two datasets, here we use distinct analytics of \textit{genre}, \textit{topic}, and \textit{persuasion-techniques} distributions across the two corpora, following the distinctions proposed for the SemEval-2023 Task 3, and using the public APIs of the news classifier GATE Cloud.\footnote{\url{https://cloud.gate.ac.uk/shopfront\#tagged=Misinformation}} 

\paragraph{Genres.} A three-fold genre distribution into Reporting articles / Opinion articles / Satire-like articles, is shown in Figure~\ref{fig:genres} for the two corpora. 

The comparison reveals significant differences between the datasets: \mainstream\ is characterized by a larger proportion of Reporting articles, more than six times the proportion in \ppn. \mainstream\ also shows a lower proportion of Satire. While these articles do not necessarily display humorous content characteristic of satire (see \citealt{icard-etal-2024-multi-label}, where the label ``satirical'', defined as intending to produce laughter, was applied less than 5\% even for propagandist articles), this finding supports previous observations that fake news are stylistically closer to satire in style than regular news~\cite{Horne_Adali_2017}.

Finally, while Opinion is significantly represented across the two corpora, including the \mainstream\ one, the class represents more than three quarters of the propagandist corpus \ppn, confirming the link between propaganda and persuasion, and the tendency of propaganda to blur the frontier between factual reports and opinion pieces.

\begin{figure*}
    \centering
    \includegraphics[width=\linewidth]{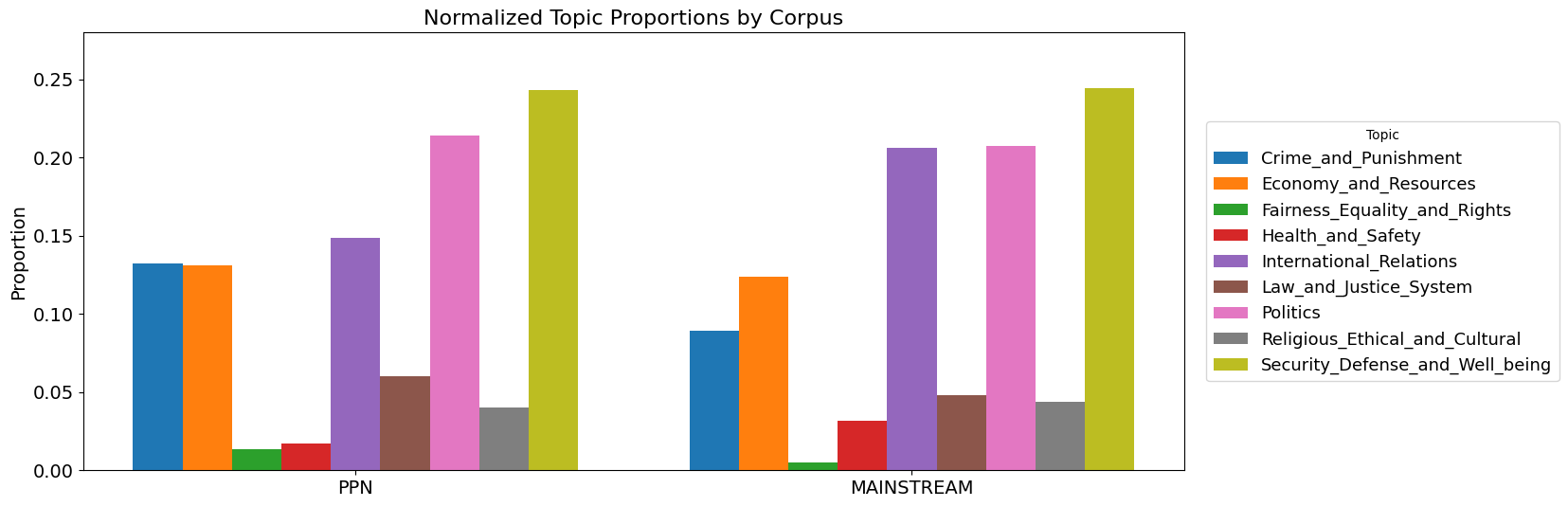}
    \caption{Topic distribution on the two corpora.}
    \label{fig:topics}
\end{figure*}

\paragraph{Topics.} We used the same suite of annotating tools to get the topic distribution of the articles. A division along nine topics is shown in Figure~\ref{fig:topics}, showing the two corpora to have relatively similar distributions.\footnote{A similar distribution was also found in the OBSINFOX dataset mentioned in fn. \ref{fn:obsinfox}.} This suggests that the two datasets can meaningfully be compared in terms of stylistic features, since they broadly have the same coverage.

\begin{figure}[!ht]
	\centering
	\includegraphics[width=0.9\linewidth]{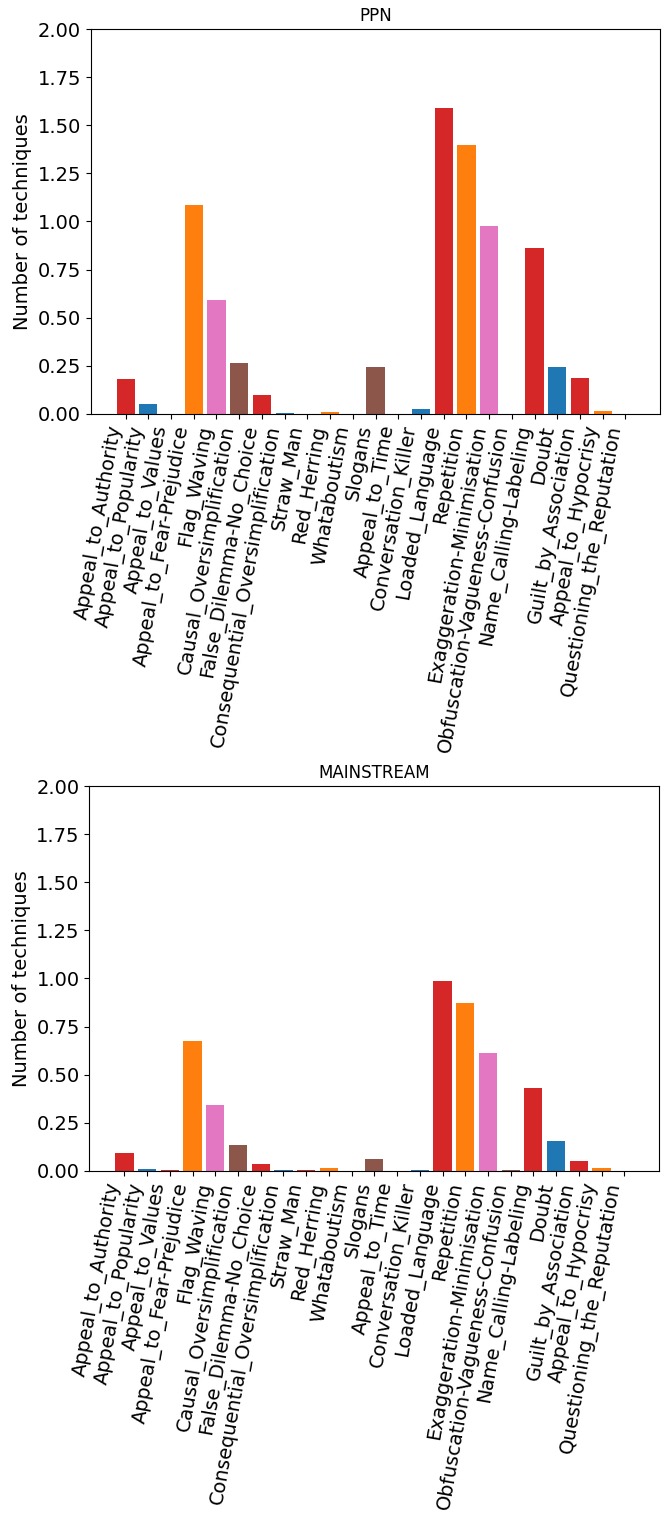}
	\caption{Mean number of persuasion techniques by article, on the two corpora.}
	\label{fig:persuasion}
\end{figure}

\paragraph{Persuasion techniques.} Finally, we used a third distributional analysis, this time relative to the set of persuasion techniques defined in~\citet{piskorski-etal-2023-semeval}, again using the Cloud multilingual persuasion technique classifier. The distribution of persuasion techniques by articles is shown in Figure~\ref{fig:persuasion}.

The plots indicate that propaganda articles from the \ppn\ corpus tend to use more of these persuasion techniques, which is coherent with the fact that more than 90\% of that corpus is identified as Opinion or Satire. In particular, we see a more prevalent use of \textit{loaded language}, \textit{repetition}, \textit{exaggeration-minimization}, and \textit{appeal to prejudice} in the propagandist corpus. 

Overall, these analyses show us that in terms of genres as well as persuasion techniques, propagandist articles are more easily recognizable than other kinds of articles. This fact, combined with the additional characteristics explored in \citet{faye-etal-2024-exposing}, suggests that we may enhance the detection of propaganda by taking into account genres, persuasion techniques, and other stylistic features. To address this, the next section introduces a hybrid approach that integrates neural and symbolic representations by combining text-based features with concepts extracted from the content.

\section{Neurosymbolic approach for propaganda detection}
\label{sec:detection}

For the remaining of the paper, we focus on the English part of the corpus (3219 articles for \texttt{PPN} and 1004 articles for \texttt{MAINSTREAM}), as the models used perform better on this language, and will allow for better quantitative evaluation of our approach. The imbalance between the classes is managed by using the \texttt{WeightedRandomSampler} of PyTorch to expose the model to a balanced amount of classes during training.

To enhance 
robustness, here we embed text using neurosymbolic methods, and we add conceptual information to the articles to enhance classification. Text embeddings are an efficient way of encoding texts and classifying them in downstream tasks. However, there is little understanding of the information they contain. To create a more robust model, we propose combining simple statistical text embeddings with the features observed in the previous section.
Thus, texts are encoded using pre-trained {fastText} embeddings~\cite{bojanowski2017enriching}. While these are non-contextual embeddings, they provide stable and parcimonious lexical representations for large amounts of data.

\subsection{Proposed architecture}

For each article, the process creates a 300-dimensional vector representing the text's distributional lexical characteristics. In addition to this vector, information about the genre, topic, and persuasion techniques contained in the articles is added.

\begin{itemize}
    \item Genre information is encoded using one-hot encoding (OHE), creating a vector containing only zeros with the exception of the encoded feature, which contains a one. In this case, it adds 3 specific dimensions (for Reporting, Opinion, and Satire).
    \item Topic information is also one-hot encoded, adding 9 dimensions (for the topics displayed in Figure~\ref{fig:topics}).
    \item Information about persuasion techniques is added into a vector counting how many persuasion techniques of each type are contained in the article. Fine-grained persuasion techniques represent 23 dimensions. However, these techniques can be grouped into coarser-grained groups, resulting in only 6 dimensions~\cite{piskorski-etal-2023-semeval}.
\end{itemize}

\begin{figure}[H]
	\centering
	\includegraphics[width=0.75\linewidth]{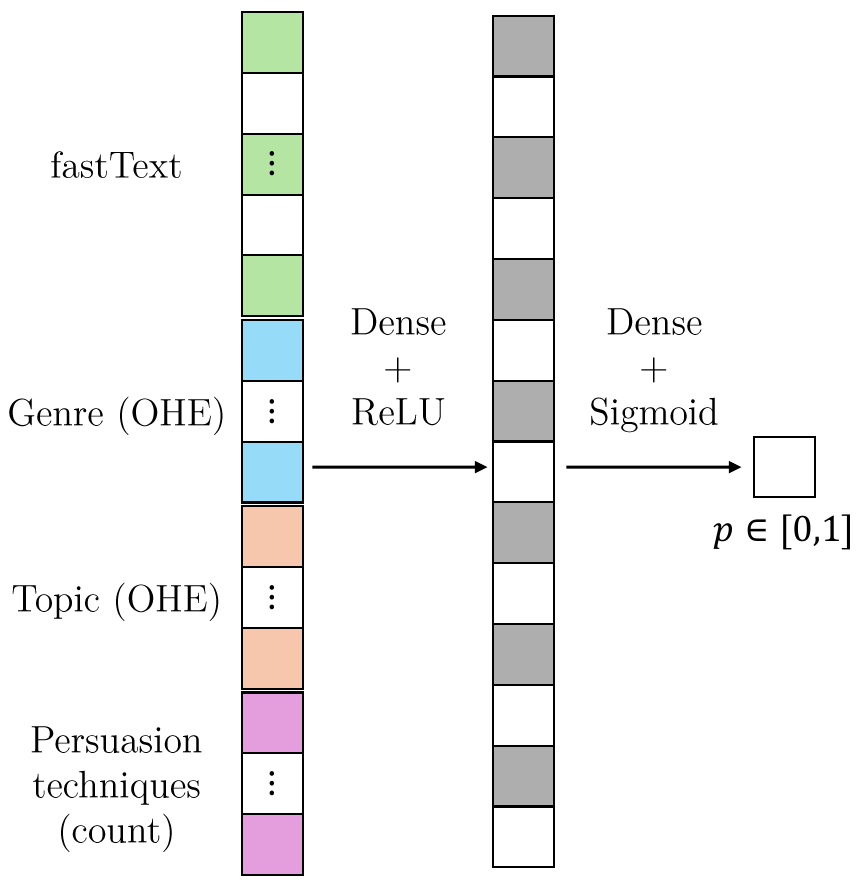}
	\caption{Our hybrid architecture, combining neural features and extracted concepts.}
	\label{fig:archi}
\end{figure}

In total, a 35-dimensional vector (=3+9+23) for fine-grained persuasion techniques is added to the 300-dimensional fastText embedding. This vector goes through a two-layer perceptron containing a dense layer (335 dimensions to 335 dimensions) with ReLU activation function, and a dense layer (335 dimensions to 1 dimension) with sigmoid activation function to get a propaganda estimation score between 0 and 1. The text is then classified relative to a threshold of 0.5. A global view of the proposed architecture is presented in Figure~\ref{fig:archi}. When all persuasion techniques are included, we call the resulting method the \textit{Hybrid Method}. A scaled-down method is obtained by using an 18-dimensional (=3+9+6) vector incorporating only coarse-grained persuasion techniques: we call the corresponding method \textit{Hybrid Lite}.

This architecture is voluntarily made simple in order to perform explainability analyses, which are hardly possible with Large Language Models. For this binary classification class, we use a single neuron as an output as we frame the task as we oppose directly reliable news to propagandist news.

\subsection{Evaluation methodology}


In order to prevent overfitting, we designed a new evaluation methodology to ensure that the proposed approach could generalize better to new events and sources.

The general idea is to split the data into train/valid/test sets depending on different criteria explained below. The model is trained on the train set and evaluated on the validation set after each epoch. Early stopping with patience 20 is used to monitor the F1-score on the validation set. The F1-score is 
the harmonic mean between precision (true positive/true positives+false positives) and recall (true positives/true positives+false negatives).
In what follows, propaganda articles are the positive class to detect.



The models are trained with a cross-entropy loss and the Adam optimizer~\cite{kingma2017adammethodstochasticoptimization} with a learning rate of $10^{-4}$ for a maximum of 300 epochs. If no improvement in the validation F1-score is observed after 20 epochs, the best model is restored and evaluated on the test set, giving the scores reported in the results tables. We ran the experiment five times with different seeds and report the average results over these 5 runs.

\begin{table*}[!ht]
  \begin{minipage}{\textwidth}
    \centering
    \footnotesize

    \resizebox{0.96\textwidth}{!}{%
    \begin{tabular}{|c|c|c|p{3.05cm}|}
      \hline
      \cellcolor{gray!15}\textbf{Sources} & \cellcolor{gray!15}\textbf{Train} & \cellcolor{gray!15}\textbf{Validation} & \cellcolor{gray!15}\textbf{Test}\\
      \hline
      \mainstream & APNews & CNN \hspace{0.5em} USA Today \hspace{0.5em} Forbes \hspace{0.5em} Fox News & CBSNews \\
                  & The Guardian & NBC News \hspace{0.5em} NYTimes \hspace{0.5em} Washington Post & Daily Mail \\
      \hline
      \ppn & RRN & \multicolumn{2}{c|}{TribunalUkraine \hspace{0.5em} War on Fakes}\\
      \hline
    \end{tabular}%
    }

    \smallskip

    \resizebox{0.96\textwidth}{!}{%
    \begin{tabular}{|c|c|c|p{3.4cm}|}
      \hline
      \cellcolor{gray!15}\textbf{Political} & \cellcolor{gray!15}\textbf{Train} & \cellcolor{gray!15}\textbf{Validation} & \cellcolor{gray!15}\textbf{Test}\\
      \hline
      \multirow{2}{*}{\mainstream}
      & APNews \hspace{0.5em} CNN \hspace{0.5em} USA Today \hspace{0.5em} NBC News
      & \multicolumn{2}{c|}{\multirow{2}{*}{Daily Mail \hspace{0.5em} Forbes \hspace{0.5em} Fox News}} \\
      & NYTimes \hspace{0.5em} Washington Post \hspace{0.5em} CBSNews
      & \multicolumn{2}{c|}{} \\
      \hline
      \ppn & \multicolumn{3}{c|}{Entire \ppn\ corpus (English)} \\
      \hline
    \end{tabular}%
    }

    \smallskip

    \resizebox{0.96\textwidth}{!}{%
    \begin{tabular}{|c|c|c|p{1.5cm}|}
      \hline
      \cellcolor{gray!15}\textbf{Credibility} & \cellcolor{gray!15}\textbf{Train} & \cellcolor{gray!15}\textbf{Validation} & \cellcolor{gray!15}\textbf{Test}\\
      \hline
      \multirow{2}{*}{\mainstream}
      & APNews \hspace{0.5em} CNN \hspace{0.5em} USA Today \hspace{0.5em} Forbes \hspace{0.5em} The Guardian
      & \multicolumn{2}{c|}{\multirow{2}{*}{Daily Mail \hspace{0.5em} Fox News}} \\
      & NYTimes \hspace{0.5em} Washington Post \hspace{0.5em} CBSNews
      & \multicolumn{2}{c|}{} \\
      \hline
      \ppn & \multicolumn{3}{c|}{Entire \ppn\ corpus (English)} \\
      \hline
    \end{tabular}%
    }

    \caption{News distribution in the \textbf{Sources}, \textbf{Political}, and \textbf{Credibility} splits.}
    \label{tab:splits}
  \end{minipage}
\end{table*}

The way the train/valid/test sets are defined can help measure robustness, namely how effective the learned features are in new contexts. Toward that goal, we define four types of split for our experiments (again on the English part of each corpus):

\begin{itemize}
    \item \textbf{Random}: The articles are randomly sampled to produce 80\%-10\%-10\% random sets.
    \item \textbf{Sources}: The split sets only contain articles from specific sources. The sources were chosen to have an approximately 80\%-10\%-10\% split distribution. The split of sources is given in Table~\ref{tab:splits} (top).
    \item \textbf{Political}: Mainstream articles are split according to their political leaning annotation by MediaBiasFactCheck.\footnote{It is important to note that MediaBiasFactCheck seems to follow the American Overton window, 
    so their political annotation may differ from what could be considered in other countries.} As a majority of articles come from left-leaning sources, we use them as training sources, and randomly split right-leaning sources between validation and test. Propaganda articles are randomly chosen for each set following an 80\%-10\%-10\% distribution (see Table \ref{tab:splits} middle). 
    \item \textbf{Credibility}: Similarly to political leaning, MediaBiasFactCheck proposes credibility ratings of sources based on freedom of the press, articles' factuality, ownership, and previous fact-checks. In this sense, all propaganda articles come from low credibility sources. As a large majority of regular articles come from high-credibility sources, we use these sources for training and low-credibility sources for validation and test. Propaganda articles are randomly chosen for each set following an 80\%-10\%-10\% distribution (see Table \ref{tab:splits}, bottom).
\end{itemize}


Each split is designed to evaluate one type of potential bias of our model based on sources, which may combine several types of unidentified biases, and then on political orientation, and credibility.

\section{Results}
\label{sec:results}

This section is divided into three parts. The first provides the results of the proposed approach on the different splits. In the second part, ablation studies are conducted to measure the benefit of adding conceptual embeddings to the textual embeddings. Finally, an explainability analysis highlights in which cases the proposed approach has more benefits than others.

\subsection{Main results}

Results for the different splits are shown in the first row of Table~\ref{tab:ablation} (Hybrid), reporting Accuracy (Proportion Correct) and F1 score. Note that the test sets are different in each column. 
The \textbf{Random} column corresponds to classical evaluation. The \textbf{Sources} column corresponds to the system being confronted with new sources, the \textbf{Political} column to the system being confronted with new political ideas, and the \textbf{Credibility} column to the system being confronted to sources of different credibility from the training set.


The results obtained on the \textbf{Random} set are not high but are decent for such a small model. The system shows comparable performance for the \textbf{Sources} and \textbf{Credibility} splits, but has more difficulties dealing with new \textbf{Political} orientations. Compared to \textbf{Credibility}, \textbf{Political} does not include \textit{Forbes} in the validation and test sets, but has it in the train set along with \textit{The Guardian} and without \textit{CBSNews}. These shifts suffice to lower performance, suggesting that the political orientation of training sets should be variegated to create more robust systems. 


\subsection{Ablation studies}

The motivation for the proposed approach is to improve robustness in new scenarios by combining conceptual features with text embeddings to reduce overfitting. To evaluate the performance of our Hybrid method, we conducted ablation studies. To begin with, a model using only the fastText embeddings is trained and evaluated (Table \ref{tab:ablation}, Text Only). Then, the features of the persuasion techniques are altered to take only into account the coarse persuasion categories (6 instead of 23, see Table~\ref{tab:ablation}, Hybrid Lite).

\begin{table*}[t]
	\centering\small
	\begin{tabular}{|c|rl|rl|rl|rl|}
        \hline\small
        & \multicolumn{2}{c|}{\textbf{Random}} & \multicolumn{2}{c|}{\textbf{Sources}} & \multicolumn{2}{c|}{\textbf{Political}} & \multicolumn{2}{c|}{\textbf{Credibility}} \\
        \cline{2-9}
		Method & Acc. & F1 & Acc. & F1 & Acc. & F1 & Acc. & F1 \\
		\hline
        Hybrid & 78.08 & 87.5 & \textbf{79.45} & \textbf{88.37} & 69.86 & 81.66 & \textbf{79.45} & \textbf{86.95}  \\
		Hybrid Lite & \textbf{79.45} & \textbf{88.54} & 34.24 & 31.42 & 67.12 & 79.66 & 56.16 & 69.81 \\
        Text Only & \textbf{79.45} & \textbf{88.54} & 20.54 & 0.0 & \textbf{79.45} & \textbf{88.54} & 20.54 & 0.0 \\
        \hline
	\end{tabular}
	\caption{Results for propaganda detection with different data splits and different ablations.}
    \label{tab:ablation}
\end{table*}

Several observations can be made:

\begin{itemize}

    \item In nearly all cases, using fine-grained labels for persuasion techniques (Hybrid) improves performance over using coarse-grained labels (Hybrid Lite). One exception is the \textbf{Random} split, but the gain is large where the Hybrid Lite method struggles (\textbf{Sources}, \textbf{Credibility}), and the loss small otherwise. 

    \item On average, the Hybrid method improves performance compared to the Text Only method (+26.89\% accuracy and +41.85\% F1-score average), even though it performs less well in \textbf{Random} and \textbf{Political}. Overall, however, the Hybrid method is the most robust across the four splits: it learns in all four cases, and it has the largest average performance with the least variance ($\mu_{F1}=86.12, var_{F1}=9.18$).
    
    
    \item By contrast, whereas the Text Only method outperforms the others in two splits (\textbf{Random}, \textbf{Political}, in the \textbf{Sources} and \textbf{Credibility} splits textual embeddings were not sufficient to learn to discriminate between propaganda and mainstream articles. This indicates that the Text Only method is not robust to perturbations of the training set, even though it does not overfit on political orientation.
    
\end{itemize}

In summary, even if the proposed approach does not perform best on traditional random splits, it shows better robustness and generalization than the equivalent text-only approach, which collapses in two cases.
Another advantage of this approach is its simplicity, allowing for the application of explainability methods, to which we turn next.

\subsection{Explainability analyses}

To explain our hybrid model, we used SHAP~\cite{SHAP_2017_NIPS}. This game-theory-based approach performs local explanations on each sample by determining which features are the most important for the final prediction. However, the produced explanations are local and dataset-dependent, they do not explain the model's behavior more generally.

\begin{figure}[ht]
	\centering
	\includegraphics[width=1.0\linewidth]{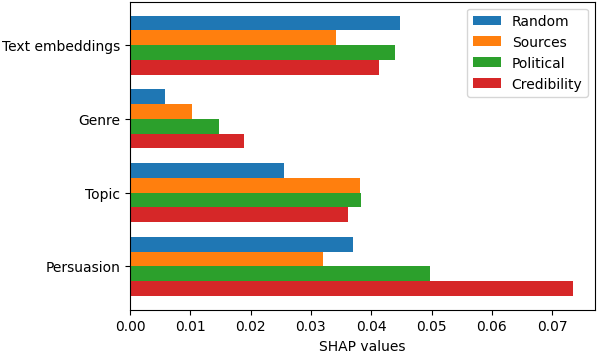}
    \vspace{-1em}
	\caption{Mean SHAP values for the Hybrid method depending on split.}
	\label{fig:shap}
\end{figure}

To get a global representation of what a model has learned, we can average absolute SHAP values over the different splits. 
We grouped SHAP values by categories for better readability. For each sample of each split, we calculate the absolute value of the sum of all text-encoding SHAP values, and similarly for the embeddings of genre, topic, and persuasion techniques. Mean SHAP values by split group and category are shown in Figure~\ref{fig:shap}.

The figure shows that for \textbf{Random}, the text embeddings contribute the most to the decision classification. For \textbf{Credibility}, the persuasion embeddings contribute comparatively the most. For \textbf{Sources} and \textbf{Political}, the results are more mixed across groups of features.

In the \textbf{Random} case, this is likely due to the fact that by choosing articles randomly, the training set is better aligned with validation and test, so they are informative enough for classification. This is confirmed by the results, which show that they are better with textual features only.

However, when the test distribution is different from the training distribution, the method tends to use at least as many conceptual embeddings when compared with text embeddings. 
In particular, for the \textbf{Credibility} and \textbf{Sources} splits, relying on conceptual features is what makes the model generalize better to new sources and to papers whose reliability is questionable. 
This situation corresponds to the case in which an article is suspicious and comes from unknown sources, making this approach suitable for propaganda detection.

\section{Conclusion}
\label{sec:conclusion}

This paper introduced a propaganda detection method that integrates textual embeddings with conceptual features extracted through cross-comparison of \texttt{PPN} and \mainstream. The corpora differ significantly in vocabulary and persuasion strategies, suggesting that models trained on a single source may miss corpus-specific signals. These observations motivated the inclusion of conceptual features to better capture propaganda patterns.

By designing biased splits in our datasets, which correspond to the exposure of the model to new types of articles, we have shown that adding conceptual information extracted from the texts improves detection performance, especially in cases where there are new sources of variable credibility ratings. Experiments also suggest that political diversity in the training set is essential for propaganda detection, as the addition of conceptual features significantly degrades performance in this case.

Further explainability analyses show that the added features were indeed used by the model when the splits were biased, allowing the model to correctly detect propaganda when simple textual embeddings are not informative enough.

However, the experiments were run on a corpus centered on one main theme:
the Russia-Ukraine conflict. Additional experiments could be conducted on other recent themes, such as recent elections, or other conflicts. The \texttt{PPN} and \mainstream\ corpora were also only processed in English, and similar experiments should be conducted in other languages to identify potential language-specific differences.

Finally, other types of conceptual features could be used based on other expert knowledge systems or even human operators. In other experiments, an expert vagueness estimation system was successfully combined with a language model for the task of subjectivity detection~\cite{clef-checkthat:2024:task2:hybrinfox}. It may be possible to add an estimate of the document's source reliability to a classification model, to enhance the classification performance of a text-only classifier.

\section*{Acknowledgments}

We thank two anonymous reviewers for helpful comments and feedback. This work was supported by the programs HYBRINFOX (ANR-21-ASIA-0003) and TRUSTEDNEWS (ANR-25-ASM2-0003). PE thanks the Department EEE of the University of Melbourne, and the Department of Philosophy of Monash University, for additional support.

\section{References}\label{sec:reference}

\bibliographystyle{lrec2026-natbib}
\bibliography{lrec2026-example}

\label{lr:ref}
\bibliographystylelanguageresource{lrec2026-natbib}
\bibliographylanguageresource{languageresource}

\end{document}